# A Rhetorical Analysis Approach to Natural Language Processing

Benjamin Englard
Dr. Michael M. Krop Senior High School
Miami, FL 33179
Email: benjienglard@yahoo.com

**Abstract**

The goal of this research was to find a way to extend the capabilities of computers through the processing of language in a more human way, and present applications which demonstrate the power of this method. This research presents a novel approach, Rhetorical Analysis, to solving problems in Natural Language Processing (NLP). The main benefit of Rhetorical Analysis, as opposed to previous approaches, is that it does not require the accumulation of large sets of training data, but can be used to solve a multitude of problems within the field of NLP.

The NLP problems investigated with Rhetorical Analysis were the Author Identification problem – predicting the author of a piece of text based on its rhetorical strategies, Election Prediction – predicting the winner of a presidential candidate's re-election campaign based on rhetorical strategies within that president's inaugural address, Natural Language Generation – having a computer produce text containing rhetorical strategies, and Document Summarization.

The results of this research indicate that an Author Identification system based on Rhetorical Analysis could predict the correct author 100% of the time, that a re-election predictor based on Rhetorical Analysis could predict the correct winner of a re-election campaign 55% of the time, that a Natural Language Generation system based on Rhetorical Analysis could output text with up to 87.3% similarity to Shakespeare in style, and that a Document Summarization system based on Rhetorical Analysis could extract highly relevant sentences. Overall, this study demonstrated that Rhetorical Analysis could be a useful approach to solving problems in NLP.

# 1 Introduction

In the past, the field of Natural Language Processing (NLP) has been primarily concerned with such applications as voice recognition, information retrieval, and machine translation, as demonstrated by Apple's development of the iPhone personal assistant, Siri, IBM's development of the Jeopardy playing computer system, Watson, and Google's development of Google Translate. These innovations have all been made possible because of the approach to Natural Language Processing known as the Maximum Entropy approach, a method which is founded upon identifying the statistical model which maximizes the inherent uncertainty of a problem [1].

The problem with a Maximum Entropy approach to the analysis of natural language, however, is that it often requires the accumulation of large sets of training data. This analysis of language, therefore, goes against the natural way in which humans analyze a piece of writing. Humans can extract information and meaning from a single piece of writing without the need for large sets of training data. One can argue that a true analysis of a sophisticated piece of writing requires a reader to perhaps have background knowledge on a topic, to perhaps have a well-developed vocabulary, but if the true goal of an Artificial Intelligence agent is to mimic and eventually surpass the capabilities of the human mind, then a computer should also be able to analyze writing without the need for the accumulation of training data. The goal of this research therefore, was to find a way to extend the capabilities of computers through the processing of language in a more natural way, and present applications which demonstrate the power of this novel method.

# 2 Rhetorical Analysis of Natural Language

This research paper presents a novel approach, Rhetorical Analysis, to solving problems in NLP. Rhetorical Analysis is an age-old practice in the human form of analysis of natural language, and involves investigating the techniques and strategies used by an author to make a piece of writing meaningful, purposeful, and memorable. Rhetorical Analysis is therefore most often used in the analysis of speeches and essays, pieces of writing which were intended to persuade an audience towards a certain end. The process of Rhetorical Analysis was begun by Aristotle, and expanded upon by the famous orator of the Roman Empire, Marcus Tullius Cicero.

This research paper will detail the benefits of a Rhetorical Analysis of natural language by first describing the method of Rhetorical Analysis, and then presenting novel solutions to several traditional NLP problems. This paper also describes many directions for future NLP research based on a Rhetorical Analysis approach.

## 3 Identifying Rhetorical Strategies

There exist a plethora of rhetorical strategies, devices used by authors and speakers to make their writing achieve a purpose and seem memorable [2]. Some rhetorical devices are easier to notice than others; however, they are all written or spoken to achieve a sense of meaning and memorability.

### 3.1 Chosen Rhetorical Strategies

For this research, a subset of the set of rhetorical strategies was chosen to be analyzed. First analyzed were the dash and the semi-colon. Although small and normally insignificant, the dash and the semi-colon represent the pauses made by an author or speaker. Since the pauses a speaker makes can often be as important as the words being uttered, the dash and the semi-colon can have a major effect on a piece of writing's memorability. In addition, writers often include dashes and semi-colons to emphasize the phrase that comes after the pause, which can thereby affect the effect of a piece of writing.

Next, three strategies were chosen for their effect on the structure of a piece of writing: alliteration, anaphora, and epistrophe. Alliteration is the starting of consecutive words with the same letter and can make a piece of writing sound either euphonious or cacophonous, eloquent or confusing. Anaphora is the starting of consecutive sentences with the same word or phrase for emphasis. The most memorable example of anaphora is Dr. Martin Luther King Jr.'s repeating of "I Have A Dream" during his famous March on Washington speech. Epistrophe is the ending of consecutive sentences with the same word or phrase and has the same effects on a reader or listener as anaphora.

The final strategy chosen for analysis was parallelism. Parallelism is the repeating of a grammatical structure in consecutive phrases and can achieve a major effect of memorability. Parallelism can be identified by analyzing the parts-of-speech of consecutive phrases of a sentence. For example, in Abraham Lincoln's "Gettysburg Address", Lincoln remarks "…of the people, by the people, for the people…" An analysis of the parts-of-speech of this phrase demonstrate the parallelism: "of, by, for" are all prepositions, "the, the, the" are all articles, and "people, people, people" are all plural nouns. In Lincoln's example, there is parallelism between three-word phrases. For this research, parallelism was identified between consecutive words, and phrases with two, three, and four words.

**3.2 Implementing Rhetorical Strategy Finders**

Once the subset of rhetorical strategies that were to be used was chosen, the process began of implementing algorithms which could find those strategies in a piece of text. Six finder programs were implemented using Python version 2.7.2. The finder programs had several common functions, shown in the pseudocode in Figure 1, with each program given helper functions as needed. Each finder program implemented a pattern-matching algorithm specific to its rhetorical strategy, and then returned a counter, representing the number of instances of each rhetorical strategy found in a piece of text.

The identification of dashes and semi-colons was completed using Python's built-in pattern-matching functions. However, in order to identify alliteration, anaphora, epistrophe, and parallelism, more advanced functionality was needed than was provided by Python's built-in libraries. Specifically, sentence tokenizers and part-of-speech taggers were needed. To allow for these functionalities, the open source Python library for Natural Language Processing, NLTK version 2.0 [3] was used. The alliteration, anaphora, epistrophe, and parallelism algorithms all used NLTK's sentence tokenizer to split a sentence into its words and punctuation. When the parallelism finder was implemented, it utilized NLTK's access to the Brown News Corpus to train a unigram part-of-speech tagger. The pseudocode algorithm for the identification of dashes is shown in Figure 2 as an example of the rhetorical strategy finder programs.

**3.3 Unittesting Finder Programs**

Once the six finder programs were implemented, each was tested on the Python unittesting platform. The programs were unittested by being given a set of sample text files with a known number of rhetorical strategies, and checking to ensure that the correct number of rhetorical strategies was found in these files. The dash (five), semi-colon (nine), alliteration (eight), anaphora (five), and epistrophe (nine) finders successfully passed all of their unittests. The parallelism finder passed only eight out of nine unittests. This is due to the fact that despite recent advancements, part-of-speech taggers still cannot identify the parts-of-speech of all words correctly.

```
class Finder:
    #Path is inputted from Unittest/Controller
    #Returns counter
    procedure count(path_to_file):
        f = file @ path_to_file
        counter = 0
        for line in f:
            counter += get_all_instances(line)
        return counter
    #Returns the length of an array which contains all
    #instances of strategy
    procedure get_all_instances(line):
        strategy_instances = []
        while more_strategies_to_be_found:
            # Add instances of strategy to array strategy_instances
        return len(strategy_instances) # length of array = counter
```

**Figure 1 – Pseudocode for Rhetorical Strategy Finder Class** (specific helper methods for each strategy's pattern-matching algorithm are not shown)

```
#Gets next occurrence of a dash
#Returns True/False value, position of dash
  procedure get_next_target(line):
      pos_dash = line.find(' - ')
      if no_dashes_in_line:
          return False, -1
      return True, pos_dash+1
#Returns the length of an array which contains all
#instances of dashes
  procedure get_all_dashes(line):
      dashes = []
      while True:
          is_dashes_in_line, position_dash = get_next_target(line)
          if is_dashes_in_line:
              dashes.append(position_dash)
              line = line[position_dash:end]
          else:
              break
      return len(dashes) # => length of dashes array = number of dashes in piece of text
```

**Figure 2 – Pseudocode for Pattern-Matching Algorithm for Finding Dashes**

## 4 Author Identification Problem

### 4.1 The Problem

The first problem that could potentially be solved using a Rhetorical Analysis of natural language and thereby without the accumulation of large sets of training data is the Author Identification Problem. The Author Identification Problem can be defined as the problem of figuring out the author of a piece of writing, given pieces of writing known to be written by that author, and pieces of writing known to not be written by that author. Previous approaches to the Author Identification Problem have involved the development of extensive classification algorithms, and the training of those algorithms with extensive sets of training data.

Rhetorical Analysis allows for a novel, more human-like solution to the Author Identification Problem. Rather than analyzing characteristics of writing as the average number of words per sentence or average number of sentences in a paragraph as previous classification algorithms have in an attempt to gain knowledge about an author's style, performing a Rhetorical Analysis allows a predictive algorithm to gain a real picture of the techniques used by an author – a method closer to what humans would use to distinguish authors. The foundational idea of the computer-generated prediction of authors through Rhetorical Analysis is the same, however, as previous approaches, in that both assume that authors maintain a specific style of writing throughout their careers, which is near impossible to duplicate.

**4.2 The Initial Algorithm**

The algorithm developed for the predicting of authors through Rhetorical Analysis relies on the similarities and differences in the usage of rhetorical strategies between authors. The algorithm begins by creating a SQLite database using Python's built-in sqlite3 library in which data obtained on the strategies used by authors will be stored. The algorithm then looks to its two-dimensional array, which holds an author's name in its first column, and the names of stored text files containing that author's piece of writing in its second column. The algorithm then loops through each row in the two-dimensional array, opens up the file, sends the file into each rhetorical strategy finder, and adds each instance of strategy counter returned by the pattern-matching algorithm to an array of counters. The algorithm then divides the counter for each strategy by the total number of strategies found to determine the relative probability of each strategy in that piece of writing, and enters the probability data combined with the author's name into a row in the SQLite table. The table created by analyzing the strategies of the initial unittesting text files is shown in Figure 3 below. The program, MesaSQLite, created the graphical user interface for the table.

| rowid | author | pDash | pSemi | pAllit | pAna | pEpi | pPara |
|---|---|---|---|---|---|---|---|
| 1 | Dash | 5.0e+1 | 0.0 | 0.0 | 0.0 | 0.0 | 5.0e+1 |
| 2 | Semi | 0.0 | 5.2941176470588204e+1 | 1.1764705882352899e+1 | 0.0 | 0.0 | 3.5294117647058798e+1 |
| 3 | Allit | 0.0 | 0.0 | 8.75e+1 | 0.0 | 0.0 | 1.25e+1 |
| 4 | Anaphora | 0.0 | 0.0 | 1.66666666666667e+1 | 5.0e+1 | 0.0 | 3.33333333333333e+1 |
| 5 | Epistrophe | 0.0 | 0.0 | 9.0909090909090899 | 2.7272727272727298e+1 | 2.7272727272727298e+1 | 3.63636363636364002e+1 |
| 6 | Parallelism | 0.0 | 0.0 | 0.0 | 0.0 | 0.0 | 1.0e+2 |
| 7 | Unknown | 1.11111111111111e+1 | 3.33333333333333e+1 | 0.0 | 0.0 | 0.0 | 5.55555555555556e+1 |

**Figure 3 – SQLite Table of Probabilities of Rhetorical Strategies for Test Files**

**4.3 Ranking Authors**

As shown in Figure 3, the table created has one row with an author named "Unknown". The task of the remainder of the algorithm is therefore to determine which of the six other authors has written the piece of writing with the probabilities of strategies shown in Row 7 in Figure 3, by analyzing the similarities in the probabilities of strategies. Similarities in the probabilities of strategies can be quantified by computing a distance measure, namely the Root-Mean-Square (RMS) deviation, between the probabilities of strategies in the unknown piece of writing and the others, six others in the case of Figure 3. If $n_s$ = number of strategies analyzed, and $P_{a,s}$ = Author *a*'s Probability of Strategy *s*, then the RMS error between the piece of writing of the unknown author and the piece of writing by author *a* is: $RMS_a = \sqrt{\left(\frac{1}{n_s}\right)\left[\sum_{s=1}^{n_s}(P_{a,s} - P_{unknown,s})^2\right]}$.

The author whom the algorithm will believe wrote the unknown piece of writing is therefore the author with the smallest RMS error, the author whose writing style was closest to that of the unknown author's. But the algorithm not only returns the best author prediction, it returns an array of authors sorted in increasing order of RMS error, or decreasing order of 100 minus the Normalized RMS error (Normalized RMS Error = $\frac{RMS_a}{RMS_{max} - RMS_{min}}$, expressed as a percent). In summary, the solution to the Author Identification Problem through Rhetorical Analysis is to rank authors in terms of smallest RMS deviation to the unknown author's piece of writing. The algorithm was then unittested to determine its accuracy as it was hypothesized to predict authors correctly 75% of the time.

## 5 Predicting Elections

### 5.1 The Problem

Recent research in the area of predicting elections has focused on using global search trends and social media with the idea that counting the number of times a political candidate is mentioned on the web can be a good predictor of an election result [4], [5]. These past studies have seen mixed results. This research attempted to find a correlation between the rhetorical strategies used in the inaugural addresses of new presidents of the United States and the results of their re-election campaigns.

### 5.2 The Solution

This problem was solved using a method similar to that of Author Identification, by comparing the relative similarities of pieces of writing using computer-generated Rhetorical Analysis of natural language in Python. First, databases containing the probabilities of strategies of presidents who gave an inaugural address and won, and presidents who gave an inaugural address and lost re-election had to be created. This was accomplished by accessing the C-Span corpus of Inaugural Addresses through NLTK. A set of 14 inaugural addresses of winners and losers were used in the creation of the database, and the results are shown in Figures 4 and 5. Once the two SQLite tables representing the probabilities of strategies in the inaugural address of presidents who have won and lost re-elections were assembled (Figures 4 and 5), the data was averaged and put into two new tables: winnersAverage and losersAverage (not shown here).

| rowid | author | pDash | pSemi | pAllit | pAna | pEpi | pPara |
|---|---|---|---|---|---|---|---|
| 1 | Washington | 8.54700854700855e-1 | 6.83760683760684 | 6.9230769230769198e+1 | 8.54700854700855e-1 | 0.0 | 2.22222222222222e+1 |
| 2 | Jefferson | 2.09424083769634 | 1.20418848167539e+1 | 5.8638743455497398e+1 | 3.1413612565445002 | 1.04712041884817 | 2.30366492146597e+1 |
| 3 | Madison | 1.8181818181818199 | 1.4545454545454501e+1 | 6.2727272727272698e+1 | 9.0909090909090895e-1 | 9.0909090909090895e-1 | 1.9090909090909101e+1 |
| 4 | Monroe | 1.0526315789473699 | 4.5614035087719298 | 7.0877192982456094e+1 | 4.9122807017543897 | 1.7543859649122799 | 1.6842105263157901e+1 |
| 5 | Jackson | 2.12765957446809 | 3.1914893617021298 | 6.3829787234042598e+1 | 0.0 | 0.0 | 3.0851063829787201e+1 |
| 6 | Lincoln | 2.3972602739725999 | 4.4520547945205502 | 5.8561643835616401e+1 | 5.8219178082191796 | 3.4246575342465801 | 2.5342465753424701e+1 |
| 7 | Grant | 5.0 | 4.0 | 6.6e+1 | 0.0 | 0.0 | 2.4e+1 |
| 8 | Cleveland | 2.3668639053254399 | 3.55029585798817 | 6.4497041420118293e+1 | 2.3668639053254399 | 5.9171597633136097e-1 | 2.6627218934911198e+1 |
| 9 | McKinley | 2.4767801857585101 | 3.09597523219814 | 6.7182662538699702e+1 | 2.4767801857585101 | 0.0 | 2.4767801857585098e+1 |
| 10 | Wilson | 0.0 | 6.0606060606060597 | 6.4848484848484802e+1 | 7.8787878787878798 | 1.2121212121212099 | 2.0e+1 |
| 11 | Roosevelt | 2.5641025641025599 | 8.2051282051282008 | 6.1538461538461497e+1 | 5.6410256410256396 | 0.0 | 2.2051282051282101e+1 |
| 12 | Roosevelt | 5.1612903225806503 | 8.3870967741935498 | 6.0645161290322598e+1 | 8.3870967741935498 | 1.93548387096774 | 1.54838709677419e+1 |
| 13 | Roosevelt | 1.6129032258064498e+1 | 2.5806451612903198 | 4.6451612903225801e+1 | 8.3870967741935498 | 1.93548387096774 | 2.45161290322581e+1 |
| 14 | Eisenhower | 5.9574468085106398 | 2.5531914893617 | 6.3404255319148902e+1 | 5.5319148936170199 | 8.5106382978723405e-1 | 2.1702127659574501e+1 |

**Figure 4 – SQLite Table of Probabilities of Rhetorical Strategies of 14 Presidents who gave an inaugural address and were re-elected.**

| rowid | author | pDash | pSemi | pAllit | pAna | pEpi | pPara |
|---|---|---|---|---|---|---|---|
| 1 | Adams | 4.9504950495049499e-1 | 8.9108910891089099 | 5.6435643564356397e+1 | 1.98019801980198 | 4.9504950495049499e-1 | 3.1683168316831701e+1 |
| 2 | Adams | 1.5625 | 1.09375e+1 | 6.5e+1 | 2.5 | 3.125e-1 | 1.96875e+1 |
| 3 | VanBuren | 2.1021021021021 | 1.08108108108108e+1 | 6.1261261261261303e+1 | 1.8018018018018001 | 0.0 | 2.4024024024024001e+1 |
| 4 | Polk | 1.5695067264574001 | 1.79372197309417 | 7.1748878923766796e+1 | 4.4843049327354301 | 1.12107623318386 | 1.9282511210762301e+1 |
| 5 | Taylor | 1.0416666666666701 | 6.25 | 7.5e+1 | 0.0 | 0.0 | 1.77083333333333e+1 |
| 6 | Pierce | 1.6025641025641 | 2.5641025641025599 | 6.5384615384615401e+1 | 3.8461538461538498 | 3.2051282051281998e-1 | 2.6282051282051299e+1 |
| 7 | Buchanan | 1.26050420168067 | 4.2016806722689098e-1 | 7.2689075630252105e+1 | 0.0 | 4.2016806722689098e-1 | 2.5210084033613398e+1 |
| 8 | Hayes | 4.2056074766355103 | 3.7383177570093502 | 6.3551401869158902e+1 | 2.3364485981308398 | 9.3457943925235e-1 | 2.5233644859813101e+1 |
| 9 | Garfield | 1.3761467889908301 | 2.75229357798165 | 7.2477064220183493e+1 | 4.5871559633027497 | 4.5871559633027498e-1 | 1.8348623853210999e+1 |
| 10 | Harrison | 2.6595744680851102e-1 | 1.59574468085106 | 7.0744680851063805e+1 | 5.8510638297872299 | 5.3194893617021705e-1 | 2.1010638297872301e+1 |
| 11 | Taft | 2.2172949002217299e-1 | 1.5521064301552101 | 7.0953436807095301e+1 | 3.7694013303769398 | 2.2172949002217299e-1 | 2.3281596452328198e+1 |
| 12 | Harding | 6.0060060060060105e-1 | 4.8048048048048004 | 5.8258258258258302e+1 | 5.7057057057057099 | 3.0030030030030003e-1 | 3.0330330330330298e+1 |
| 13 | Coolidge | 0.0 | 1.25 | 6.90625e+1 | 7.5 | 6.25e-1 | 2.15625e+1 |
| 14 | Hoover | 1.2626262626262601 | 7.8282828282828296 | 5.8585858585858603e+1 | 4.2929292929292897 | 5.0505050505050497e-1 | 2.7525252525252501e+1 |

**Figure 5 – SQLite Table of Probabilities of Rhetorical Strategies of 14 Presidents who gave an inaugural address and were not re-elected.**

The algorithm for predicting elections is then very similar to that of Author Identification: pass another inaugural address through the rhetorical strategy finders, determine the probability of each strategy, and calculate the Root-Mean-Square (RMS) error between the probabilities of strategies in the new inaugural address with the averages of winners and losers. If the RMS error between the new inaugural address and the winners is smaller than the RMS error between the new inaugural address and the losers, the algorithm assumes that this inaugural address would win a re-election campaign, and vice versa. The algorithm was then unittested to determine its accuracy and was hypothesized to predict re-elections correctly 75% of the time.

## 6 Natural Language Generation

### 6.1 Background

In 1950, mathematician Alan Turing devised his seminal "Turing Test", which argued that machines will have truly become "intelligent", when a human will not be able to discern whether it is speaking to a human or a machine during a conversation [6], [7]. Ever since then, the problem of Natural Language Generation, the problem of giving a machine the functionality to produce natural language, has been a problem of utmost importance to the fields of Artificial Intelligence, Machine Learning, Natural Language Processing, and even Neuroscience [8].

In this research project, the implementation of pattern-matching algorithms has allowed for the quantifying of an author's style of writing. The Author Identification Problem is founded on the idea that it is nearly impossible for one to replicate the style of another author. But theoretically, if a program existed which could take as input the probabilities of strategies of an author, and output text with the near exact probabilities as those inputted, that program will have effectively replicated the style of another author.

Returning to the issue of the Turing Test, it is known that rhetorical strategies are what give an author or speaker a distinct style. So if a program could be implemented which could output text with a distinct style, that program would seem more human. If the probabilities of strategies of Shakespeare were inputted, and text with the near exact probabilities as inputted were outputted and then read in conversation to a human, it would appear to the human that he or she is conversing with Shakespeare. Therefore, combining Rhetorical Analysis with Natural Language Generation could hold the key to the eventual building of machines which can pass the Turing Test.

### 6.2 The Solution

The goal of this research was to produce a Natural Language Generation system, which could take as input the probabilities of the six strategies used before, and output several sentences with the near exact probabilities inputted. The first algorithm that needed to be developed was the one that could produce sentences. For this a lexical dictionary was needed and the WordNet dictionary developed at Princeton University [9] was chosen because it can be accessed from NLTK. The algorithm developed has the user enter a first word, which is added to a sentence.

Using WordNet, the definition of that word is found and the middle word of the definition is added to a sentence. Then, the middle word from the definition of the definition of the first word is found, and is added to a sentence. The algorithm proceeds with this recursive definition until the base case is met, namely, the average number of words per sentence times the number of sentences (both values are initially inputted by a user) equals the number of words that have been written so far. This algorithm ends up producing writing with a Context-Free Grammar, writing generated by applying a set of recursive productions to an initial non-terminal, until a final goal is reached [10].

The next goal of the Natural Language Generation system was to develop a method by which rhetorical strategies could be inputted into text, beginning with dashes and semi-colons. Using the initial probabilities entered by the user and the number of sentences to write entered by the user, the algorithm randomly selects which sentences to input dashes and semicolons into, so that the near exact probabilities as inputted will occur in the output text. When the algorithm reaches the sentence at which it knows a strategy will be inputted, it inputs the dash or semi-colon at a randomly chosen position in the sentence.

Later work involved adding the functionality to the Natural Language Generation system to write text with a distribution of sentence lengths to give the system a more natural style, since humans do not write the same number of words in each sentence.

After this initial development, it was found that the program did not need to be developed any further, because alliteration, anaphora, epistrophe, and parallelism were already being inputted due to the relationships between words and their definitions. For example, when "bird" is inputted, the first three words of output are: "Bird characterized character", which contains alliteration.

In order to test how similar the output text was to Shakespeare, the probabilities of each strategy in Shakespeare's *Hamlet*, *Macbeth*, and *Julius Caesar* were found using the pattern-matching algorithms. Six different files were generated by the Natural Language Generation system using the probabilities found in Shakespeare, and their Normalized RMS errors from Shakespeare's plays were calculated. Since a Normalized RMS error would equal the overall difference in style between writings, 100 minus the normalized error would equal the overall similarity as a percent. It was hypothesized that on average, the computer-generated texts would be 75% similar to Shakespeare's writings.

Since the algorithm was not developed to input alliteration, anaphora, epistrophe, and parallelism into the text, and since it was found that Shakespeare used no dashes in a majority of the writings used, rather than input the actual probabilities of Shakespeare when generating texts, the algorithm had inputs of an average of 10 words per sentence, five sentences to be written, and 100% semicolons.

**6.3 Randomizing the Algorithm**

The main problem of the original algorithm was that given a single input word, the algorithm always outputted the same text. However, there is no strict grammar to human language; we are essentially free to write any word after any word. This is especially true for authors like Shakespeare who are free to use "poetic license", implying they are less likely to follow the rules of English grammar when writing. Because of this, it was believed that a randomized algorithm for producing text would end up being closer in style to human produced text. So, the algorithm was randomized by having it choose which word from the definition of an input word to add to the sentence being written randomly. This ends up producing a probabilistic automata, and a Markov Chain, where, if $P(w_n)$ equals the probability of word n, $P(w_n) = \left( \frac{1}{\#words-in-def'n-of:w_{n-1}} \right)(P(w_{n-1}))$. Interestingly, $P(w_n)$ is defined recursively, just as the algorithm is. Because it was not known how randomizing the algorithm would affect the probability of strategies within the output text, it was hypothesized that the 12 randomly computer-generated texts would be 75% similar to Shakespeare's writings – the same as the previous estimate.

**6.4 Calculating Entropy**

When the original tests of the Natural Language Generation system were completed, it was found that as believed, the output of the randomized algorithm was closer in style to the Shakespearean text than the output of the original algorithm. This would imply that Shakespeare's text is more random than non-random, as it was closer in style to the output text of the randomized algorithm.

Entropy is a measure of the randomness of a system, and in 1948, Claude Shannon developed a method to calculate the entropy of writing [11]. Shannon's entropy (H) is defined as $H = -\sum_i p_i \log_2 p_i$, where $p_i$ equals the probability of state $i$. The entropy of language can be calculated therefore, by adding up the products of the probability of each word and the log base two of the probability of each word. The relative entropy of the writing can be calculated by dividing H by the maximum entropy, the maximum entropy equaling $-\log_2(\#distinct-words)$. Shannon estimated the relative entropy of the English language to be 50%, implying that during our communication, we follow the rules of grammar only half of the time [11].

Since the randomized text was closer in style to the Shakespearean text, it was hypothesized that the entropies of the pieces of writing would be similar. Specifically, it was hypothesized that the relative entropy of the output text of the non-random algorithm would be 75%, the relative entropy of the output text of the random algorithm would be 95%, and the relative entropy of the Shakespearean text would be 85%. It was also hypothesized that higher entropies would imply higher percent similarities, as calculated in section six of this paper.

# 7 Document Summarization

## 7.1 Background

Computer-generated document summarization is a traditional NLP problem, dating back to the 1950's [12], [13], and involves developing an algorithm which can extract the most important sentences from a piece of text. Past solutions include analyzing word frequencies, and then creating a summary by extracting the sentences which contain the most of the highest frequency words [12]. This, however, is a purely statistical approach and disregards the linguistic features which make a sentence seem important. Summarization is a difficult problem because the computer-generated summary must contain background information, the author's purpose, and the main idea – linguistic features which are hard to notice without some sort of Natural Language Understanding system.

## 7.2 Rhetorical Analysis Approach

Rhetorical Analysis allows for a novel solution to the problem of Document Summarization. Often, the sentences in speeches which humans would consider the most important, are the ones they can remember. As examples, most would probably agree that "…this nation, under God, shall have a new birth of freedom -- and that government of the people, by the people, for the people, shall not perish from the earth" is the most important sentence of Abraham Lincoln's Gettysburg Address or that "…ask not what your country can do for you—ask what you can do for your country" is the most important sentence of former President John F. Kennedy (JFK)'s inaugural address. These sentences are so memorable because they contain rhetorical strategies.

Therefore, if one wanted to extract the most important sentences from a document for summary, one must find the sentences with the most rhetorical strategies.

### 7.3 The Solution

Using the rhetorical strategy finder programs, the rhetorical strategies within a sentence could be found and their probabilities calculated. In order to extract the most important sentences for summary, a method of scoring and ranking sentences had to be developed. If $\omega_s$ = the weight of strategy $s$, $n_{sentence}$ = the number of strategies found in *sentence*, and $P(s|sentence)$ = the probability of strategy $s$ within *sentence*, then the score $S$ of a sentence can be calculated as: $S(sentence) = \frac{n_{sentence}}{6}\sum_s \omega_s P(s|sentence)$. For dashes and semi-colons, $\omega_s$ = 0.05 each, for alliteration $\omega_s$ = 0.1, for anaphora and epistrophe $\omega_s$ = 0.2 each, and for parallelism, $\omega_s$ = 0.4. These weights were determined based on how they affect the memorability of a sentence: parallelism can have the greatest effect and was therefore given the largest weighting, whereas dashes and semi-colons have the least effect and were therefore given the smallest weighting. The sum is then multiplied by the average number of each strategy in the sentence, in order to ensure that sentences with more rhetorical strategies were ranked highest. Sentences are then sorted according to score and the four sentences with highest score are outputted.

## 8 Results and Discussion

### 8.1 Code

In total, nearly 2500 lines of Python and SQL (Structured Query Language) code were written for this research representing six rhetorical strategy finder programs and their unittests, the author identifier, the election predictor, the language generation system, the entropy calculator system, and the document summarizer. All of the algorithms described in sections three, four, five, six, and seven were implemented successfully.

### 8.2 Author Identification Problem Results and Discussion

Using NLTK's access to the works in the Project Gutenberg [14] corpus, the Author Identification system was tested to determine its accuracy. The chosen texts came from such notable authors as William Shakespeare, John Milton, Herman Melville, Lewis Carroll, Jane Austin, G.K. Chesterton, Maria Edgeworth, William Blake, Walt Whitman, Thornton W. Burgess, and Sara Cone Bryant. In addition, excerpts were taken from Dr. Martin Luther King Jr.'s "I Have a Dream" speech and Abraham Lincoln's "Gettysburg Address" for analysis.

Tests were conducted by sending in pieces of text written by the authors listed above, and then adding another piece of text which was known by the researcher to have been written by one of the authors in the list, but which was unknown by the computer. The goal was to have the computer predict the author of the unknown piece of writing correctly 75% of the time, using the algorithm described in section four. The data shown in Table 1 represents the results of the tests. The extra author column corresponds to the number of additional authors used in the tests, besides the one unknown text, and the one author who was known by the researcher to have written the unknown piece of writing.

| # Extra Authors | # Tests Ran | % Tests Correct |
|---|---|---|
| 1 | 6 | 100.0 |
| 2 | 5 | 100.0 |
| 3 | 4 | 100.0 |
| 4 | 4 | 100.0 |
| 5 | 5 | 100.0 |
| 6 | 4 | 100.0 |
| 7 | 4 | 100.0 |
| 8 | 4 | 100.0 |
| 9 | 4 | 100.0 |
| 10 | 4 | 100.0 |
|  | Total: 44 | Average: 100 |

**Table 1 – Results of Author Identification Problem Accuracy Tests**

Overall, the Author Identification system predicted the correct author of an unknown piece of writing with 100% accuracy, surpassing the hypothesized 75%.

In addition to revealing that the algorithm was more successful than it was initially believed to be, the results of the tests often revealed interesting literary trends. For example, in every test where the algorithm should have predicted Shakespeare as the author of an unknown piece of writing, it did so correctly. This demonstrates that Shakespeare maintained a very unique style throughout his writings which no one could duplicate successfully, which is in fact the foundation of the Author Identification problem. What is more interesting is that the two authors most similar to Shakespeare in style were William Blake and Walt Whitman. The pieces of text written by Blake and Whitman, which were used in the tests, were poems, which implies that Shakespeare wrote in a poetic style even in his plays.

Furthermore, the author closest in style to Jane Austen was Maria Edgeworth. This is interesting because Jane Austen and Maria Edgeworth are two of the only three female authors among the list of authors chosen for analysis. This implies that rhetorical analysis may be able to distinguish the gender of an author, which could be a useful tool in plagiarism detectors, as well as text sentiment analyzers.

**8.3 Election Predictor Results and Discussion**

When the winnersAverage and losersAverage tables were created, it was apparent that the probabilities of strategies in the two tables were extremely similar. Specifically, across the six strategies, there was an average standard deviation between the probability of strategies of the winnersAverage and losersAverage of only 0.7998. This hinted that interestingly, presidents throughout history have used very similar style in their inaugural addresses, and that because of this similarity, the election predictor would not be as reliable as initially hypothesized.

For the 38 presidents who delivered an inaugural address and ran a re-election campaign, the algorithm correctly predicted the results of the re-election 21 times, or about 55% of the time. This was lower than the original hypothesis of 75%. However, while a rhetorical analysis approach to predicting elections did not prove to be as reliable as expected, it did provide insight into our past presidents. As previously stated, the probability of strategies within the inaugural address of re-election campaign winners and losers were extremely close, implying that our past presidents have used very similar styles when delivering their first speech as president. This would imply that despite the events of the time, or the major campaign issue, presidents generally use the same techniques when speaking, in order to give their speech a sense of purpose, meaning, and memorability.

As another metric for the extreme similarity between the probabilities of strategies of the re-election winners and losers, the standard deviation of the RMS errors of each inaugural address when compared to the winnersAverage and losersAverage was calculated. It was found that across all of the tests of the election predictor, the average standard deviation was 0.5639, further indicating the similarities in probabilities of strategies between the re-election winners and losers.

For the November 2012 election for president, the algorithm predicted that Barack Obama would win his re-election campaign. The standard deviation between the RMS errors was 0.937, implying a greater difference in the RMS errors than the average, implying a greater certainty of the result. Using the algorithms for predicting authors, it was also found that President Obama's 2009 inaugural address was most similar in style to the 1937 inaugural address given by Franklin Delano Roosevelt (FDR). Interestingly, this could possibly have been predicted even without mathematical analysis, as both presidents spoke of similar themes, namely, for people to remain hopeful in times of economic depression.

Overall, rhetorical analysis did not prove to be as useful as expected in the problem of predicting the results of re-elections. However, it did reveal an interesting insight into the nature of inaugural addresses.

**8.4 Natural Language Generation Results and Discussion**

Using the initial non-randomized algorithm, and the input words of bird, generalization, hand, hasty, indeed, and passion, six texts were outputted. The output texts included: "Bird characterized character work making results is an associate in ; equal", "Hand extremity part relation ; characteristic attribute or United . Unite in ; equal another various but nothing quantity there here . Present continuous time", and "Indeed often times period amount quantity". Interestingly, without being programmed to do so, the algorithm ended up producing writing with other rhetorical strategies. This would imply a deep relationship between words and their definitions.

Using the second algorithm, two randomized texts were outputted for each of the original six input words. Here, the output texts included more variety, and some examples include: "Bird egg envelopes usually conditions process achieve gain .", "Hasty quick is copula as arsenic rat rodents single on operational . Process intended as compounds union an in length in length linear first . Series things ; especially is an degree quality attribute . Can paint liquid pressure in length place region ; extended . Or on or ; in length dimension width side relative marriage married .", "Hasty quick toenail ; toe digits collectively conjunction things movable or . On operational achieving gain is have ; possesses force effect phenomenon ."

By calculating the normalized RMS errors of each outputted text compared to the three chosen Shakespearean plays, the efficiency of the Natural Language Generation experiment could be determined. The efficiency would imply, expressed as a percent, how close the outputted text was in probabilities of strategies to the inputted Shakespearean style text. It was initially estimated that for both the non-randomized and the randomized algorithms, 100 minus the normalized RMS errors would be about 75%, implying that the outputted texts would be 75% similar to Shakespeare in style. The results are indicated in the table below. All texts are random, unless otherwise indicated.

|  | % Similarity *Hamlet* | % Similarity *Macbeth* | % Similarity *Julius Caesar* |
|---|---|---|---|
| Bird 1 (non-random) | 76.7 | 74.1 | 75.7 |
| Bird 2 | 78.7 | 76.5 | 77.8 |
| Bird 3 | 73.2 | 70.5 | 72.2 |
| Generalization 1 (non-random) | 77.9 | 75.4 | 76.9 |
| Generalization 2 | 80.5 | 77.3 | 79.3 |
| Generalization 3 | 84.5 | 81.6 | 83.4 |
| Hand 1 (non-random) | 80.0 | 77.7 | 79.1 |
| Hand 2 | 84.5 | 81.6 | 78.0 |
| Hand 3 | 79.2 | 76.0 | 81.7 |
| Hasty 1 (non-random) | 81.6 | 79.1 | 80.6 |
| Hasty 2 | 82.7 | 79.6 | 81.7 |
| Hasty 3 | 87.3 | 84.9 | 86.3 |
| Indeed 1 (non-random) | 83.1 | 80.7 | 82.2 |
| Indeed 2 | 80.9 | 78.0 | 79.8 |
| Indeed 3 | 86.3 | 83.4 | 85.2 |
| Passion 1 (non-random) | 79.2 | 76.9 | 78.4 |
| Passion 2 | 82.3 | 79.3 | 81.2 |
| Passion 3 | 80.6 | 77.8 | 79.7 |
| Average Non-Randomized Results | 79.7 | 77.3 | 78.8 |
| Average Randomized Results | 81.7 | 78.9 | 81.1 |

**Table 2 – Results of Natural Language Generation Percent Similarity Tests**

Of the 54 tests shown above, 50/54 or about 93% of normalized RMS similarities were over 75%. Of the 18 tests of the non-randomized algorithm, 17/18 or about 94% of normalized RMS similarities were over 75%. Of the 36 tests of the randomized algorithm, 33/36 or about 92% of normalized RMS similarities were over 75%. Overall, the average normalized RMS similarity of the non-randomized algorithm's output texts was 79.7%. This surpassed the hypothesis, that the average normalized RMS similarity of the non-randomized output text would be 75%. Overall, the average normalized RMS similarity of the randomized algorithm's output texts was 81.1%.

This surpassed the hypothesis, that the average normalized RMS similarity of the randomized output text would be 75%. In addition, it validated the initial belief that the randomized text would be closer in style to Shakespeare than the non-randomized text because of Shakespeare's ability to use "poetic license" when writing.

**8.5 Entropy Results and Discussion**

The first hypothesis of the relationship of entropy to the Natural Language Generation system was that the relative entropy of the output text of the non-random algorithm would be 75%, the relative entropy of the output text of the random algorithm would be 95%, and the relative entropy of the Shakespearean text would be 85%. When the actual entropies were calculated, the results were as follows:

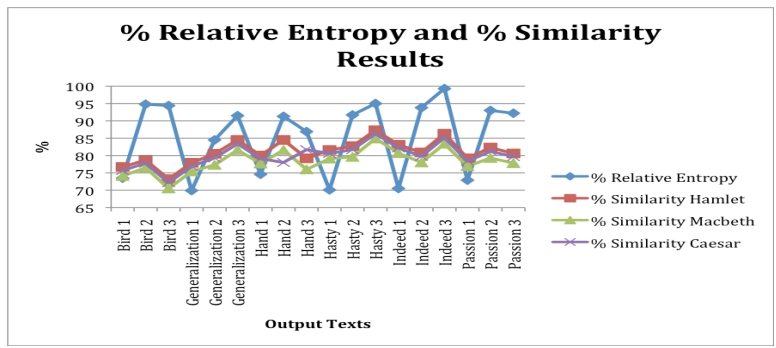

**Figure 6 – Results of Comparisons of Relative Entropies with % Similarities**

The data from Figure 6 clearly demonstrates that as expected, the relative entropies of the output texts from the randomized algorimthms were higher than the output texts from the non-randomized algorithm. The data also indicated that the average relative entropy of the output text of the non-randomized algorithm was 71.9%, lower than the hypothesized 75%, the average relative entropy of the output text of the randomized algorithm was 92.4%, lower than the hypothesized 95%, the average relative entropy of the Shakesperean text was 66.9%, lower than the hypothesized 85%. The reason why Shakespeare's entropy was lower than the entropy of the output text of both the non-random and randomized algorithms, was that his maximum entropy was higher because he used a larger vocabulary – a larger number of distinct words.

It was also hypothesized that higher entropies would imply higher percent similarities, as calculated in section six of this paper. The results demonstrated in Figure 6 clearly validate this belief. In this graph, the peaks and troughs of the curves occur at the same places along the x-axis, indicating that increases in relative entropy lead to increases in percent similarities, and vice versa.

These results demonstrated that although the relative entropy of Shakespeare was less than that of the randomized or even non-randomized algorithm, randomizing the algorithm seems to have affected the output text in a way that brought the probabilties of rhetorical strategies closer to that of Shakespeare's. In this sense, perhaps the most important hypothesis: that adding rhetorical strategies to a Natural Language Generation system could make the system seem more human and in turn help machines pass the Turing Test, seems to have been validated. In addition, it appears that the key to solving the Turing Test will be in how machines are programmed to produce language more randomly so as to mimic human style, rather than how they are taught, for example, to follow the rules of English grammar.

**8.6 Document Summarization Results**

Since no metrics were developed which could quantify the success of the Document Summarization system, speeches were chosen for which the researcher knew the most important sentences and could analyze the output. The speeches chosen were: Abraham Lincoln's "Gettysburg Address", Dr. Martin Luther King Jr.'s "I Have a Dream" speech, former President JFK's inaugural address, President Barack Obama's inaugural address, and FDR's 1933 inaugural address. Future work could determine a system by which the success of the Document Summarization system could be quantified.

As previously stated, the difficulty of Document Summarization is the task of identifying sentences which convey background information, the author's purpose, and the main idea, without a system which can attempt to understand the meaning of the text. An analysis of the summaries generated by the Rhetorical Analysis based Document Summarizer demonstrates that all of those linguistic features are captured in the four highest ranked sentences. For example, the three highest ranked sentences (ie. the summary) of Dr. Martin Luther King Jr.'s "I Have a Dream" speech were: "Five score years ago, a great American, in whose symbolic shadow we stand today, signed the Emancipation Proclamation. And when this happens, when we allow freedom ring, when we let it ring from every village and every hamlet, from every state and every city, we will be able to speed up that day when all of God's children, black men and white men, Jews and Gentiles, Protestants and Catholics, will be able to join hands and sing in the words of the old Negro spiritual: Free at last. This is no time to engage in the luxury of cooling off or to take the tranquilizing drug of gradualism." In the summary, the first sentence represents background information, the second – main idea, and the third – author's purpose. Therefore, it can be concluded that the Rhetorical Analysis approach produced output just as well as a Natural Language Understanding system could; because, rather than analyzing purely statistical measures, the summarizer described analyzes the techniques the author or speaker used – just as a human would.

## 9 Conclusion and Future Research

Once the results of the study were analyzed, the next question was: "has Rhetorical Analysis proven itself useful?" Given the results of the tests of the Author Identification System, the Natural Language Generation system, the Entropy Calculator, and the Document Summarizer, Rhetorical Analysis appears to be an effective and novel method. Given the results of the tests of the Election Predictor, Rhetorical Analysis appears to be only a somewhat useful method.

However, all of this research was limited by the fact that only six strategies were analyzed and used in solving problems. Future research will involve implementing pattern-matching algorithms for more rhetorical strategies. This will involve current research in sentiment analysis, so that strategies such as verbal irony and apophasis, which rely on the tone of the writing, can be identified. With more strategies to quantify an author's style, the accuracy of the results of problems that can be solved with Rhetorical Analysis can only increase.

Future research will also involve recent advances in machine translation. The pattern-matching algorithms which find dashes and semi-colons in a piece of text, will theoretically work with any language which contains those symbols. The alliteration finder will work with any languages for which the first character of consecutive words can be determined. The anaphora and epistrophe finders will work with any language that uses periods, exclamation points, or question marks to signal the end of a sentence. Recent development of a universal part-of-speech tagset [15], will allow for the identification of parallelism in multiple languages. The adaption of the pattern-matching algorithms to other languages will allow for the replication of this research, with a focus on pieces of writing in other languages. In addition, the extension of the Author Identification Problem to other languages will allow for the building of a global index of data on authors' styles, which could lead to better spam and plagiarism detectors, as well as become an invaluable research reference to historians and linguists around the world.

Using NLTK's access to a corpus of State of the Union Addresses, future research in the prediction of the results of re-election campaigns will be possible – this time addressing the question of whether comparing rhetorical strategies in the State of the Union Addresses of past presidents, can predict the results of future elections. By analyzing speeches delivered prior to an election, for example, speeches at each political party's national convention, Rhetorical Analysis may be able to predict the results of all elections, rather than just re-elections.

The next step in Natural Language Generation is twofold. The first step would be the further development of the system so that it can generate text without the need for a starting word inputted by a human and can respond to a human's speech. The second step would be the development of a Reverse Turing Test based on Rhetorical Analysis to investigate whether it would be possible to discern human versus computer-generated writing based on rhetorical strategies.

The branch of mathematics, which seeks to describe Markov chains as physical systems, is Ergodic Theory. According to Ergodic Theory, if an ergodic system is run for an inordinate amount of time, it will eventually reach an equidistribution of states. For the Natural Language Generation system, this would imply each word would appear in the output text with equal probability, regardless of the input word. If it could be proven, that regardless of the input word, each word within the Wordnet dictionary which contains a definition could be reached by following the non-deterministic Context-Free Grammar described in section six, then the Natural Language Generation system could be considered an ergodic system, and after being run for infinite time, each word would appear in the output text with equal probability. This would imply that each word contained in the Wordnet subset of the English language is lexically connected, which could reveal a fascinating insight into the development of language. As it stands, this could be considered a language analog of the Ergodic Hypothesis of physics [16].

This research began as an attempt to find a method by which computers could process language more naturally. The process of Rhetorical Analysis was chosen and since applications had to be found which could demonstrate the power of this novel method, this research expanded into the areas of literary analysis, the social sciences, statistics, and information theory. Using this and future research, by teaching a robot Rhetorical Analysis, that robot would theoretically be able to both analyze and write literature, and speak conversationally with a human. Therefore, given the results of this research study, one can conclude that the initial goal of this project, to extend the capabilities of computers through the processing of language in a more natural way, was achieved.

# 10 References

[1] Adam L. Berger, Vincent J. Della Pietra, and Stephen A. Della Pietra. A maximum entropy approach to natural language processing. Comput. Linguist., 22(1):39, March 1996.

[2] Robert A. Harris. A handbook of rhetorical devices. http://www.virtualsalt.com/rhetoric.htm, 2011.

[3] Steven Bird. Natural language toolkit. https://sites.google.com/site/naturallanguagetoolkit/Home, 2011.

[4] Catherine Lui, Panagiotis T. Metaxas and Eni Mustafaraj. On the predictability of the u.s. elections through search volume activity. 2011.

[5] Andranik Tumasjan, Timm O. Sprenger, Philipp G. Sandner, Isabell M. Welpe. Predicting elections with twitter: What 140 characters reveal about political sentiment. Proceedings of the Fourth International AAAI Conference on Weblogs and Social Media, pages 178-185.

[6] A. M. Turing. Computing machinery and intelligence. Mind, LIX:433-460, 1950.

[7] Brandon Keim. Artificial intelligence could be on brink of passing turing test. http://www.wired.com/wiredscience/2012/04/turing-test-revisited/, 2012.

[8] Chris Brew Jon Oberlander. Stochastic text generation. Philosophical Transactions of the Royal Society A: Mathematical, Physical and Engineering Sciences, 358:1373-1378, 2000.

[9] Princeton University. Wordnet: A lexical database for english. http://wordnet.princeton.edu/, 2012.

[10] Context-free grammars. http://www.cs.rochester.edu/~nelson/courses/csc_173/grammars/cfg.html

[11] Claude E. Shannon. A mathematical theory of communication. The Bell System Technical Journal, 27:623656, 1948.

[12] Luhn, P. H. Automatic creation of literature abstracts. IBM Journal (1958), 159-165.

[13] Jade Goldstein, Mark Kantrowitz, Vibhu Mittal, and Jaime Carbonell. Summarizing text documents: sentence selection and evaluation metrics. In Proceedings of the 22nd annual international ACM SIGIR conference on Research and development in information retrieval, SIGIR '99, pages 121-128, New York, NY, USA, 1999. ACM.

[14] Project Gutenberg. Free ebooks by project gutenberg. http://www.gutenberg.org/, 2012.

[15] Slav Petrov, Dipanjan Das, Ryan McDonald. A universal part-of-speech tagset. Proceedings of the 8th International Conference on Language Resources and Evaluation, 2012.

[16] Cesar R. de Oliveira, Thiago Werlang. Ergodic hypothesis in classical statistical mechanics. Revista Brasileira de Ensino de Fsica, 29, 2007.